\pdfoutput=1

\documentclass[11pt]{article}

\usepackage[acceptedWithA]{tacl2021v1}

\usepackage{times}
\usepackage{latexsym}

\usepackage{amsmath,amssymb}
\usepackage{graphicx}
\usepackage{cleveref}
\usepackage{enumitem}
\usepackage{booktabs}

\usepackage[T1]{fontenc}

\usepackage[utf8]{inputenc}

\usepackage{microtype}

\usepackage{inconsolata}

\newcommand{\update}[1]{#1}

\title{Navigating the Landscape of Hint Generation Research: From the Past to the Future}

\author{
Anubhav Jangra$^\diamond$\Thanks{Corresponding author - anubhav@cs.columbia.edu} \quad Jamshid Mozafari$^\dagger$ \quad Adam Jatowt$^\dagger$ \quad Smaranda Muresan$^\diamond$ \\ \ \\
    $^\diamond$Department of Computer Science \\
    Columbia University, USA  \\ \ \\
    $^\dagger$Department of Computer Science \\
    University of Innsbruck, Austria \\ \ \\
  }

\begin{document}
\maketitle
\begin{abstract}
Digital education has gained popularity in the last decade, especially after the COVID-19 pandemic. With 
the improving capabilities of large language models to reason and communicate with users, envisioning intelligent tutoring systems (ITSs) that can 
facilitate self-learning is not very far-fetched. One integral component to fulfill this vision 
is the ability to give accurate and effective feedback via hints to scaffold the learning process. In this survey article, we present a comprehensive review of prior research on hint generation,
aiming to bridge the gap between research in education and cognitive science, and research in AI and Natural Language Processing. 
Informed by our findings, we propose a formal definition of the hint generation task, and discuss the roadmap of building an effective hint generation system aligned with the formal definition, including open challenges, future directions and ethical considerations. 

\end{abstract}

\begin{figure}[!ht]
    \centering
    \includegraphics[width=\linewidth]{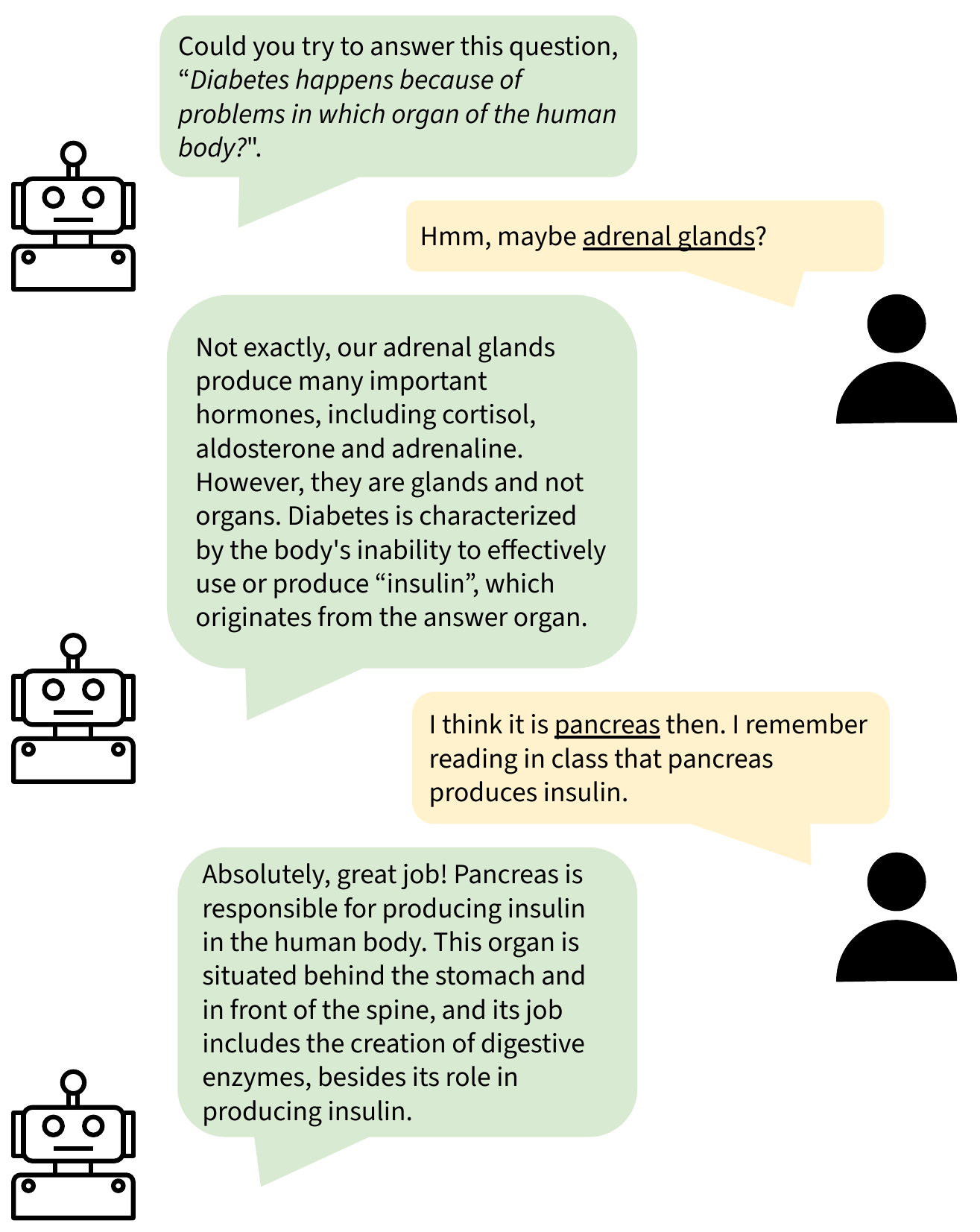}
	\caption{An example of a hint generation system capable of acknowledging the learner's wrong answer, and scaffolding them to the correct direction. 
 }
	\label{fig:intro-fig}
\end{figure}

\section{Introduction} \label{sec:intro}


Prior research has established a correlation between the student-teacher ratio and student's overall performance \cite{koc2015impact}. However, private tutoring
is not accessible to everyone, and finding expert tutors is often difficult and incurs considerable costs \cite{bray1999shadow, graesser2012intelligent}. Intelligent tutoring systems (ITSs) hold the key to addressing these educational challenges, notably the need for personalized learning in a system often reliant on instructional teaching and standardized testing \cite{anderson1985intelligent}. 

The hallmark of intelligent tutoring systems is their ability to provide step-by-step guidance to students while they work on problems, and hints play a critical role in their ability to provide this help \cite{biswas2014combined}. Hints are a tool to provide scaffolded support to the learners, and can be traced back to the socio-cultural theory of Vygotsky’s Zone of Proximal Development (ZPD), referring to "\textit{the gap between what a learner can do without assistance and what a learner can do with adult guidance or in collaboration with more capable peers}" \cite{lev1979mind}.

Within \update{learning science}, hints refer to the clues, prompts, questions, or suggestions provided to learners to aid them in solving problems, answering questions, or completing tasks, thereby encouraging critical thinking, problem-solving skills, and independent learning. In Figure \ref{fig:intro-fig}, we provide an example of a hint generation system capable of the reasoning required to answer the question, acknowledging the wrong attempt by the learner and providing informative hints linked to the learner's existing knowledge. This framework of scaffolding is well-established in education \cite{van2010scaffolding}, and expert tutors are guided to incorporate it in their teaching practices \cite{belland2017instructional}. 

With the aim of developing hint generation systems with capabilities such as the ones showcased in 
Figure \ref{fig:intro-fig}, we consolidate the dispersed efforts on hint generation, bridging the gap between research in education and cognitive sciences on the one hand, and research in AI and Natural Language Processing (NLP) on the other. Grounded in the findings from our literature review, we provide a 
roadmap for future research on automatic hint generation. 
We summarize the key characteristics of a successful hint as observed by research with human tutors in Section \ref{sec:anatomy} and review the automated hint generation systems in Section \ref{sec:computational_survey}. We identify the gaps and propose a roadmap for future research in hint generation in Section \ref{sec:roadmap}. 
We provide a rethinking of the formal definition of the hint generation task (Section \ref{sec:task_def}), a brief review of research areas that can inform the design of a hint generation system that aligns with the formal definition
(Section \ref{sec:comp_tasks}), open challenges for future directions for effective automatic hint generation systems (Section \ref{sec:future_directions}), and ethical considerations 
(Section \ref{sec:ethics}). Our major contributions include:
\vspace{-0.5em}
\begin{enumerate}[itemsep=0pt]
    \item A literature review on hint generation that bridges the gap between research in education and cognitive science on the one hand and research in AI and NLP on the other. 
    \item A formal definition of the hint generation task, grounded in the cognitive theories on learning and findings from qualitative research.
    \item A roadmap for research on automatic hint generation, outlining challenges, promising future directions, and ethical considerations for the field.
\end{enumerate}

\section{Anatomy of a Hint} \label{sec:anatomy}

In this section, we draw on research from education and cognitive sciences to describe the key characteristics of an effective hint formulation process. We start by describing the pragmatics of a hint (`context'), covering some prominent traits exhibited by expert tutors and educators while generating hints, and then dive deeper into the anatomy of a hint by discussing the \textit{semantic} (`what to say') and the \textit{stylistic} (`how to say it') aspects of a hint. 

\subsection{Pragmatics of a Hint} \label{sec:pragmatics}
Expert tutors are able to provide high quality support to students because they are aware of each student's individual learning style, strengths and weaknesses. These tutors often exhibit a contextual awareness about their students, and we describe two such common practices adopted by educators’ when supporting students below. 


\textbf{Scaffolding Support}: Structuring hints in a scaffolded manner, with incremental steps leading to the solution helps learners systematically build their understanding. These “just-in-time interventions” \cite{wood1976role} allow students to build their understanding step by step, starting with foundational concepts and progressing toward more advanced aspects without being overwhelmed by the information or task complexity \cite{zurek2014scaffolding, lin2012review, hammond2001scaffolding}. 

\textbf{Personalization and Learner Feedback}: Every learner is unique and has different needs and preferences when it comes to learning \cite{bulger2016personalized}. Prior studies point towards a learner-centered pedagogical system, where personalization and individualization of learning have a significant role in the students’ overall learning process and strengthen their sense of self and individuality \cite{radovic2012new}. To generate effective hints, it is important to recognize and cater to these individual needs by considering learners’ strengths, challenges, cultural sensitivity, and preferences
\cite{chamberlain2005recognizing, ibrahim2016assessment, suaib2017use}.
A good learning environment also incorporates a feedback loop, where hints are accompanied by opportunities for learners to provide feedback, promoting active engagement. This two-way communication allows tutors to gauge the effectiveness of their guidance and adjust their learning plans \cite{boud2013rethinking}.

\subsection{Semantics of a hint} \label{sec:semantics}

Semantics of a hint refers to the information conveyed by the hint, which includes explaining the key concepts and ideas required to scaffold the learning process.
We observed the following properties of effective hint's semantics:

\textbf{Relevance to the Learning Objective}: Learning objectives serve as a measure of achievable goals that articulate what learners should know or be able to do by the end of a learning experience. Learning objectives can broadly be categorized across three domains: cognitive, affective, and psychomotor objectives \cite{hoque2016three, sonmez2017association}. Each domain has different expectations and goals to assess the effectiveness of a hint. A hint generation system should model these objectives to create successful high-quality hints.

\textbf{Link to Prior Knowledge}: A successful hint would act as a bridge between a learner’s existing knowledge base and the current learning step to foster continuity in learning. Studies have shown that building on prior knowledge helps students bridge gaps, clear misconceptions, and reinforces the relevance of new information \cite{hailikari2008relevance, hailikari2007exploring, dong2020does}.

\textbf{Conceptual Depth}: Many learning sessions focus on teaching learners how to harness latent cognitive abilities and mold them into \textit{deep conceptual thinkers} with the ability to discuss and question more, seeking to understand rather than only memorize \cite{rillero2016deep}. It is important to balance the complexity of a hint that strikes a student’s interest without overwhelming them.

\subsection{Style of a hint} \label{sec:style}

Expert human tutors adopt diverse techniques to convey information to learners. These strategies vary from non-verbal cues such as body language, facial expressions, and vocal tone \cite{bambaeeroo2017impact, wahyuni2018power} to adopting multimedia content to teach complex topics (\textit{e.g.} using animations and maps to teach the geographical concept of "folded mountains") \cite{kapi2017multimedia}. We try to cover the most relevant aesthetic aspects of hints that might be useful in building better hint-generation system.

\textbf{Clarity and Simplicity}: Hints should be expressed in clear and simple language to ensure that learners easily grasp the underlying concept or problem-solving strategy. Avoiding unnecessary complexity enhances the usefulness of the hint and is usually well-received by the learners. This is a well-established practice within the learning sciences community known as \textit{direct instructions} \cite{kozloff1999direct, kim2005direct, rosenshine2008five}.

\textbf{Encouragement and Positive Tone}: The role of encouragement and positive attitude has been extensively investigated in several human studies in classroom settings, and all unanimously align with the significance of motivating learners towards better performance, increased participation, and improved self-confidence \cite{ducca2014positive, yuan2019teacher, li2021application, lalic2005role}. 
A hint generation systems could benefit by incorporating a positive, encouraging tone (as demonstrated in Fig. \ref{fig:intro-fig}).

\textbf{Adopting Creative and Multi-modal Elements}: In order to encourage active participation and retain learners’ interest, human tutors often adopt several creative and multi-modal elements to facilitate better understanding and information retention. These creative elements include interactive literary devices like analogy \cite{richland2015analogy, gray2021teaching, nichter2003education, thagard1992analogy}, questions \cite{hume1996hinting, chi1996constructing}, and metaphors \cite{low2008metaphor, sfard2012metaphors, guilherme2018discussing}. We can also expand beyond text, and incorporate information from other modalities such as maps \cite{winn1991learning}, diagrams \cite{winn1991learning, swidan2019role, tippett2016recent}, and multimedia content \cite{abdulrahaman2020multimedia, collins2002teaching, kapi2017multimedia} to effectively complement the learning experience. A good hint can take inspiration from some of these creative elements for a successful transfer of knowledge.


A good instructor typically takes the general guidelines into consideration and uses a mixture of the aforementioned semantic and stylistic features to create effective hints based on their prior tutoring experiences. For instance, to develop a hint that uses the literary device of analogy, the tutor must understand the prior knowledge of the learner to create successful hints \cite{gray2021teaching}.

\section{Survey of Computational Approaches} \label{sec:computational_survey}

In this section, we provide a comprehensive overview of the recent advancements of computational approaches for automatic hint generation. We first describe the extensively studied hint generation techniques for computer programming that focus on revealing code snippets to help learn how to program. Next, we dive into the \update{relatively under-explored natural language}
hint generation, where we explore 
strategies for diverse domains like mathematics, language acquisition, or factual entity-based questions. 
We conclude this section by describing some limitations of automatic hint generation systems today, and \update{propose a roadmap for future research in the field in \Cref{sec:roadmap}}.

\subsection{Hint Generation for Computer Programming}
A vast majority of computational approaches for hint generation 
have focused on the specific domain of computer programming, 
owing to the more objective nature of the task and abundance of data. We briefly discuss the approaches, datasets, and evaluation metrics adopted in the field. 
For a more comprehensive review in this specific domain, we refer the readers to the surveys written by \citet{le2013review}, \citet{crow2018intelligent}, \citet{mcbroom2021survey} and \citet{mahdaoui2022comparative}. 

\noindent\textbf{Datasets.} Two widely popular datasets in the programming hint generation space are \texttt{iSnap} \cite{price2017isnap} and \texttt{ITAP} \cite{rivers2017data} datasets. Both datasets consist of detailed logs collected from several students working on multiple programming tasks, including the complete traces of the code and records of when the hints were requested. \texttt{iSnap} \cite{price2017isnap} is based on Snap\textit{!}\footnote{\href{https://snap.berkeley.edu/}{https://snap.berkeley.edu/}} -- a block-based educational graphical programming language, while \texttt{ITAP} \cite{rivers2017data} is a Python dataset collected from two introductory programming courses taught at Carnegie Mellon University. In Table \ref{tab:examples}, we describe an example from the \texttt{ITAP} dataset, where the goal is to write a program to determine if a given day is weekend. Given a student's code that fails to pass the pre-determined test cases (\textit{e.g.,} \textit{isWeekend(``Sunday")} will return False), the aim of a hint generation system is to provide hints to help them successfully solve the problem (\textit{e.g.,} replacing lowercase `saturday' to uppercase `Saturday').

\begin{figure}[!ht]
    \centering
    \includegraphics[width=\linewidth]{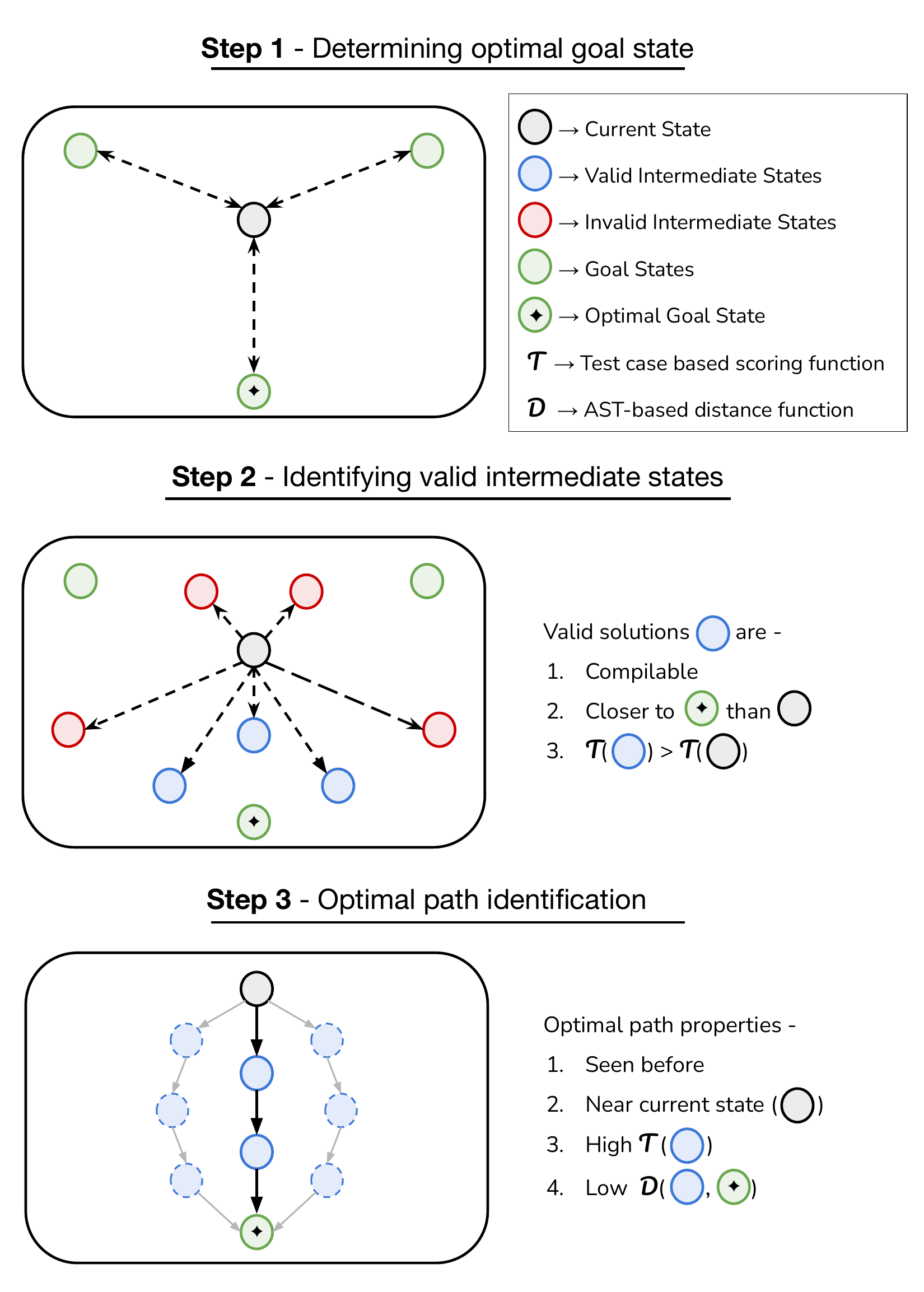}
	\caption{Illustration of the \textit{path construction algorithm} \cite{rivers2014automating}, that generates programming hints for an ongoing student attempt (current state), given reference solution(s) (goal states) and a test case-based scoring function $\mathcal{T}$.} \label{fig:programming}
\end{figure}

\noindent\textbf{Approaches.} Most of the recent efforts in programming hint generation adopt a data-driven deterministic approach \cite{barnes2008toward, rivers2017data, obermuller2021guiding, jin2012program, zimmerman2015automated, paassen2018continuous, price2016generating, rolim2017learning}, that comprises of three key components: a corpus of diverse candidate solutions (usually obtained via past student attempts), a matching algorithm to select the best candidate response given an ongoing attempt based on similarity, and graph-based solution path construction to synthesize hints (Fig \ref{fig:programming} describes one approach in detail). 
We found abstract syntax tree (ASTs) \cite{mccarthy1964formal, knuth1968semantics} to be the most popular choice of graph representation for hint synthesis due to its vast literature and language-agnostic nature. 
Similarly, \citet{mcbroom2021survey} 
provide a detailed generalization of hint generation techniques for the programming domain called \texttt{HINTS} (Hint Iteration by Narrow-down and Transformation Steps framework). Although effective, programming hints are rarely natural language responses and, therefore, are not capable of incorporating the stylistic aspects of hints that improve the learner's experience (Section \ref{sec:style}). We believe an amalgamation of NLP technologies and current hint generation systems could improve upon this limitation.

\noindent\textbf{Evaluation metrics.} \update{
Given the vast majority of work adopting a data-driven hint strategy, programming hints are designed to reveal some aspect of the program not written by the learner. Within our survey, all the evaluation metrics adopted a code-based similarity measure to gauge the quality of the hint, and we found two prominent evaluation paradigms that we categorize into \textit{reference-based} and \textit{reference-free} evaluation metrics. The \textit{reference-based} evaluation metrics \cite{price2019comparison} assume the availability of reference hints developed by expert tutors. For instance, \texttt{QualityScore} \cite{price2019comparison} is a reference-based evaluation metric that uses abstract syntax tree-based 
similarity measure to evaluate the quality of the generated programming hints with respect to expert tutor written hints. }

\update{\textit{Reference-free} evaluation metrics \cite{rivers2017data, paassen2018continuous, obermuller2021guiding, zimmerman2015automated} focus on measuring the impact of a hint to the learner's response. 
For example, \citet{paassen2018continuous} proposed root-mean-square error over two distance measures: (1) distance between the predicted post-hint state and true next state, and (2) distance between the predicted post-hint state and learner's true final state. \citet{rivers2017data}, on the other hand, adopted a learner-agnostic approach to evaluate the hint quality. They used two measures of a successful chain of hints: (1) the ability of the hint sequence to reach the correct solution state and (2) the length of the chain.}

\update{\textit{Reference-based} evaluation metrics are able to compare the quality of generated hints with the expert tutor's feedback capabilities, but can be difficult to scale. \textit{Reference-free} evaluation metrics, on the other hand, are capable of evaluating previously unseen problems but require human evaluations to obtain performance signals for the metrics. Each of these paradigms has pros and cons, but currently, there is 
no holistic measure of 
the quality of programming hints that emphasizes 
the pragmatics, semantics and style characteristics of the generated hints.
}


\begin{table*}[]
\centering
\resizebox{0.99\textwidth}{!}{%
\begin{tabular}{lll}
\toprule
\begin{tabular}[c]{@{}l@{}}\textbf{Data}\\ \textbf{Source}\end{tabular} &
  \textbf{Input} &
  \textbf{Expected or Generated outputs} \\
  \midrule
\texttt{ITAP} &
  \begin{tabular}[c]{@{}l@{}}\underline{Question}- \\ Write a program that if a given day is weekend.\\ \underline{Learner's response}- \\ def isWeekend(day): \\      \hspace{4mm} return bool(day=='sunday' or day =='saturday')\end{tabular} &
  \begin{tabular}[c]{@{}l@{}}Hint generated by \cite{rivers2017data}- \\ \textit{Type: Replace}\\ \textit{Old expression: "saturday"}\\ \textit{New expression: "Saturday"}\end{tabular} \\\hline
\texttt{ReMath} &
  \begin{tabular}[c]{@{}l@{}}\underline{Question}- \\ Mike has 4 cookies and he eats 3 cookies.\\ So Mike has \_\_\_\_ cookies left?\\ \underline{Learner's response} - \\ He has 10 cookies left.\end{tabular} &
  \begin{tabular}[c]{@{}l@{}}Expected output- \\ Error type- \textit{guess}\\ Response strategy- \textit{provide a solution strategy}\\ Response intention- \\ \textit{Help student understand the lessons topic or solution strategy}\\ Response-\\ \textit{Great try! Let's try to draw a picture.} \\ \textit{Let's start with 4 cookies and erase the 3 that Mike eats.}\end{tabular} \\\hline
\texttt{SQUAD} &
  \begin{tabular}[c]{@{}l@{}}\underline{Question}- \\ Who became the most respected entrepreneur\\ in the world according to Financial Times in 2003?\\ \underline{Expected Answer}- \\ Bill Gates\end{tabular} &
  \begin{tabular}[c]{@{}l@{}}Hint Generated by \cite{jatowt2023automatic}- \\ \textit{The searched person held the position of chief executive officer.}\end{tabular} \\\hline
\update{\texttt{TrivialQA}} & 
  \begin{tabular}[c]{@{}l@{}}\underline{Question}- \\ In which city are the headquarters of the International Monetary Fund?\\ \underline{Expected Answer}- \\ Washington D.C.\end{tabular} &
  \begin{tabular}[c]{@{}l@{}}Hint Generated by \cite{mozafari2024triviahg}-\\\textit{The city is known for its neoclassical architecture.}\\\textit{The city is located on the Potomac River.}\\\textit{The city is the capital of the USA located on the east coast.}\end{tabular} \\\hline
\end{tabular}%
}
\vspace{-0.5em}
\caption{Hint generation examples from selected prior research discussed in Section \ref{sec:computational_survey}.}
\label{tab:examples}
\end{table*}

\subsection{\update{Natural Language} Hint Generation}
\update{Natural language hint generation has risen to popularity in the last few years as a consequence of the recent advancements in NLP, particularly large language models 
capable to generate fluent and coherent text \cite{min2023recent, naveed2023comprehensive}.}
We found the question answering format to be the most prevalent setup for natural language hint generation systems, where the learner attempts to answer a question to recall and concertize their understanding of a concept. 

\noindent\textbf{Datasets.} \texttt{ReMath} \cite{wang2023step} is a benchmark \update{co-developed with math teachers (\textit{i.e.,} experts)} for evaluating and tutoring students in the mathematics domain. \texttt{ReMath} provides a systematic breakdown of the human-tutoring process into three steps: 1) identifying the error type, 2) determining a response strategy and intention, and 3) generating a feedback response that adheres to this strategy (example in Table \ref{tab:examples}). \update{Each of these steps is manually annotated by an expert math teacher.} \citet{wang2023step} also provides a set of error types (\textit{e.g.,} guess, careless, misinterpret, right-idea), response strategies (\textit{e.g.,} explain a concept, ask a question) and intentions (\textit{e.g.,} motivate the student, get the student to elaborate the answer) to facilitate the feedback generation process. \texttt{ReMath} is a great example of high-quality data collection with human experts towards building better hint generation systems. 

\update{
\texttt{TriviaHG} \cite{mozafari2024triviahg} is another hint generation dataset developed by extending the \texttt{TriviaQA} dataset \cite{joshi2017triviaqa}. \citet{mozafari2024triviahg} utilize Microsoft's \texttt{CoPilot}\footnote{\url{https://copilot.microsoft.com/}} to generate hints due to its retrieval augmented generation (RAG) approach that generated more reliable responses grounded on internet-retrieved documents. \texttt{TriviaHG} includes 10 generated hints for questions that \texttt{CoPilot} is capable on answering. In a follow-up work, \citet{mozafari2024exploring} propose \texttt{HintQA} that utilizes these hints as concise context, improving the QA capabilities of LLMs over other context-retrieval and context-generation baselines.
} 

\noindent\textbf{Approaches.} For open-ended hint generation of factoid questions, \citet{jatowt2023automatic} proposed a Wikipedia-based retrieval framework for the “Who?”, “Where?” and “When?” type questions. They propose a popularity-based framework for the "When" question type, where the popularity of an event for the answer year is measured by the count of Wikipedia hyperlinks directing to the event's website, and a hand-curated template approach for the "Who" and "Where" question types (example hint described in Table \ref{tab:examples}). On the other hand, 

\citet{wang2023step} benchmarked the \texttt{ReMath} dataset by instruction fine-tuning \cite{wei2021finetuned} the language models like \texttt{Flan-T5} \cite{chung2022scaling} and \texttt{GODEL} \cite{peng2022godel}, and using in-context learning \cite{dong2022survey} prompts for \texttt{gpt-3.5-turbo} and \texttt{gpt-4} \cite{achiam2023gpt}. \citet{tack2022ai} and \citet{wang2023step} found the direct use of LLMs to fall short in comparison to human expert responses. 

\update{\citet{pal2024autotutor} proposes a hint generation framework for middle-school level math word problems (termed \texttt{MWPTutor}). \texttt{MWPTutor} provides a hint by formulating a question around the next operation to be performed in the state space. This hint is obtained by matching the ongoing response with all possible decomposed solutions obtained by using a language model to decompose the solutions into atomic mathematical operation steps.
}
Current question answering hint generation systems do not personalize the hints to learners' preferences, learning objectives, or 
their prior knowledge (Sections \ref{sec:pragmatics} and \ref{sec:semantics}). We discuss how we can improve these hint generation systems \update{to aid the learning process} in Section \ref{sec:roadmap}.

\noindent\textbf{Evaluation Metrics.}  All the discussed approaches for 
hint generation 
have used human evaluation to assess the quality of the system's output. 
\citet{jatowt2023automatic} conducted a between-subjects study 
\update{to evaluate their proposed hint generation strategies across different experimental groups.}
\citet{tack2022ai} proposed the "AI Teacher Test", comparing the generated responses against the teacher responses across three dimensions - “\textit{speak like a teacher}”, “\textit{understand a student}”, and “\textit{help a student}”. They identified that the LLMs are good at conversation uptake (i.e., the first two requirements) but are quantifiably worse than real teachers on several pedagogical dimensions, especially \textit{helpfulness to a student}. \citet{wang2023step} evaluate the error type identification and feedback response strategy selection as a multi-class classification task and utilize exact match and Cohen’s kappa to measure the accuracy and entropy to measure the output diversity. They also conducted human evaluations for the response generation task and found that all models constrained by knowledge of ground-truth error type and response strategy outperformed their unconstrained counterparts. 

\update{
For evaluation of hint generation systems in quantitative user studies, \citet{pal2024autotutor} extends the idea of \textit{success rate} ($S$: fraction of correctly answered questions) and \textit{telling rate} ($T$: fraction of conversations where the answer was revealed) proposed by \citet{macina2023mathdial}. \citet{pal2024autotutor} suggested an \textit{adjusted success} ($S-T$) to prevent an overtly revealing framework from achieving high performance, and the harmonic mean of success rate and adjusted success rate as the overall \textit{tutoring score} (\textbf{$\frac{2S(S-T)}{2S-T}$}).
}

\update{
\cite{mozafari2024triviahg}, on the other hand, proposed two learner-agnostic automatic evaluation metrics: \textit{convergence} to measure the ability of a hint to eliminate wrong candidate answers, 
and \textit{familiarity} to measure the recognizability of answer entities. To measure \textit{convergence} of a hint, they adopt a three-step process - i) generating candidate answers using LLMs, ii) validating the entailment of answer given a hint, and iii) computing an aggregate score for a hint across all candidate answers. For \textit{familiarity}, they utilize the Page Views of Wikipedia\footnote{\url{https://www.wikipedia.org/}} articles corresponding to the named entities present in the hint as a measure of global familiarity to the hint normalized across all questions in the \texttt{TrivialHG} corpus \cite{mozafari2024triviahg}.
}

\section{Roadmap for Future Research in Hint Generation} \label{sec:roadmap}

\begin{figure*}[ht]
	\centering
	\includegraphics[width=\linewidth, scale=0.25]{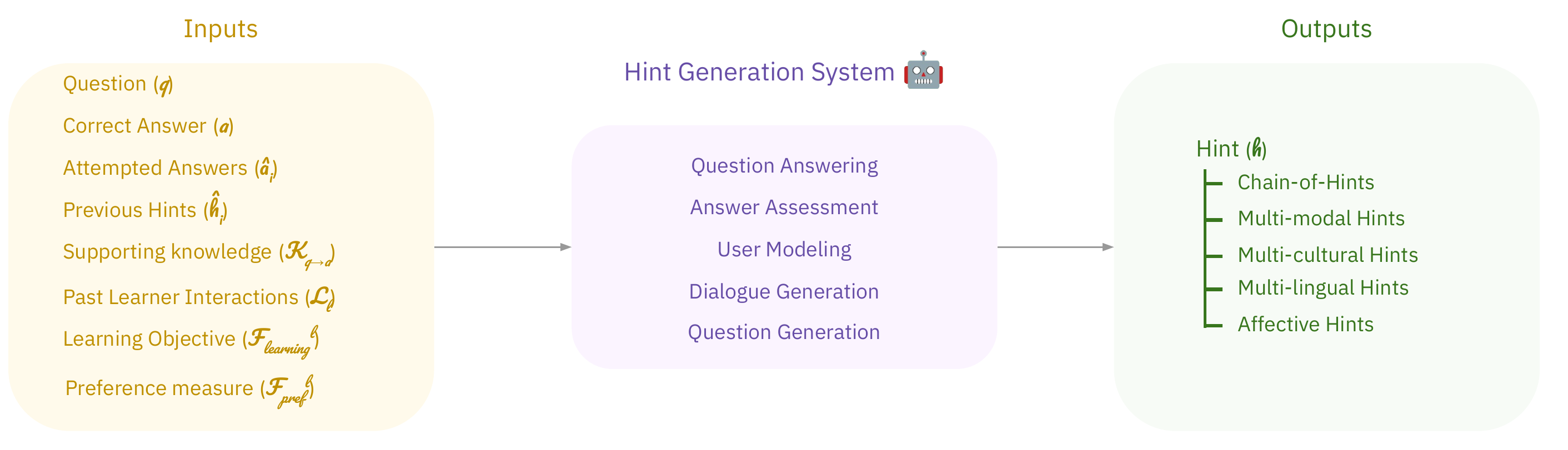}
 \vspace{-2em}
	\caption{Roadmap of the proposed \update{computational} hint generation framework.}
	\label{fig:overview}
\end{figure*}

Great progress has been made in automatic hint generation over the last two decades; however, there is still room for improvement. \update{Existing hint generation} \update{frameworks} do not personalize the hints to the learner's prior knowledge (Section \ref{sec:pragmatics}), and are only \update{evaluated} in short-term studies. We still do not know the effects of long-term exposure to these \update{interventions} on the learners. These frameworks are also limited to certain domains and have not been widely explored in other domains, including different branches of science and social science.
To improve upon these factors, we discuss a roadmap for future efforts in automatic hint generation. We propose a 
computational hint generation framework that draws on
education and cognitive sciences (Figure \ref{fig:overview}). 


\subsection{Formal Definition} \label{sec:task_def}


\citet{jatowt2023automatic} proposed a formal definition for hint generation as follows: 
Given a question $q$ and its correct answer $a$, the task is to generate a hint $h$, such that $P(a|q,h) - P(a|q) > \epsilon$, where $P(a|q,h)$ denotes the probability of a user answering $q$ if the hint $h$ is given, $P(a|q)$ is the probability of a user answering $q$ without the hint, and $\epsilon$ is a threshold parameter ($\epsilon > 0$).
This definition emphasizes 
the hint's ability to help answer a question and does not incorporate any pedagogical aspects or principles. It also does not take individual preferences into consideration.

Below, we provide a comprehensive definition that draws inspiration from widely adopted cognitive frameworks on human learning, such as Anderson's \textit{adaptive control of thought rational (ACT-R) theory} \cite{anderson2013adaptive} and Ausubel's \textit{theory on meaningful learning} \cite{ausubel1963cognitive, ausubel1962subsumption, ausubel2012acquisition}. We also incorporate the findings from the qualitative experiments discussed in Section \ref{sec:anatomy}, explicitly integrating the alignment of hints to students' learning objectives and their prior knowledge (Section \ref{sec:semantics}), as well as incorporating the pragmatics of a hint (Section \ref{sec:pragmatics}) by accounting for learners' preferences.
We formulate the hint generation task within a tutoring framework with the goal of correctly answering a question. We will later explain how the definition can be extended to generate hints for tasks beyond question answering. 

\paragraph{Refined formal definition.} 
Given a learner $l$ attempting to answer a question $q$, 
a {\bf hint generation system} $\textbf{H}$ (Figure \ref{fig:overview}) 
generates a hint $h \in \mathcal{H}$ by mapping $\textbf{H}: \mathcal{I} \rightarrow \mathcal{H}$, where
$\mathcal{I} = \{q, a, \mathcal{K}_{q\rightarrow a}, D_q^l, \mathcal{L}_l, \mathcal{F}^l_{learning}, \mathcal{F}^l_{pref}\}$ is the input to the hint generation system having the following elements:
\vspace{-0.5em}
\begin{itemize}[itemsep=0pt]
\item $q$: question
\item $a$: correct answer
\item $\mathcal{K}_{q\rightarrow a}$: supporting knowledge for the question answer pair $<q, a>$, 
\item $D_q^l = \{q, \hat{a}_1, \hat{h}_1, \hat{a}_2, \hat{h}_2, ..., \hat{a}_k\}$: ongoing dialogue, where $\hat{a}_i$ and $\hat{h}_i$ are, respectively, the learner's past attempts and hints related to $q$,
\item $\mathcal{L}_l = \{D_{q_i}\}$: learner $l$'s past learning history, 
\item $\mathcal{F}^l_{learning}$: a function to measure the learner $l$'s learning objective(s), and 
\item $\mathcal{F}^l_{pref}$: a measure of learner $l$'s preference of and familiarity to a hint.
\end{itemize}

\noindent\update{\textbf{H} generates the hint \textit{h} that} does not contain the answer (\Cref{eq:def1}), helps the learner to answer the question (\Cref{eq:def2}), and aligns with the learner's learning objective(s) (\Cref{eq:def3}).

\vspace{-0.5em}
\begin{equation} \label{eq:def1}
    P(a | q, h, D_q^l) < 1
\end{equation}
\vspace{-1em}
\begin{equation} \label{eq:def2}
    P(a | q, h, D_q^l) - P(a | q, D_q^l) > \epsilon_p
\end{equation}
\vspace{-1em}
\begin{equation} \label{eq:def3}
    \begin{split}
        \mathcal{F}^l_{learning}(q \rightarrow D_q^l \rightarrow h \rightarrow a) - \\
        \mathcal{F}^l_{learning}(q \rightarrow D_q^l \rightarrow a) > \epsilon_f
    \end{split}
\end{equation}
If the hint generation model $\textbf{H}$ is capable of generating multiple hints that satisfy the \Cref{eq:def1,eq:def2,eq:def3} such that $\textbf{H}$ returns $n$ hints $\{h_1, h_2, ..., h_n\}$ in decreasing order of preference, then -
\begin{equation} \label{eq:def4}
    \begin{split}
        (\mathcal{F}^l_{pref}(h_i) > \mathcal{F}^l_{pref}(h_j)) \rightarrow (i < j) \hspace{2mm} \\
        \forall \hspace{2mm} h_i, h_j \in \{h_1, h_2, ..., h_n\}
    \end{split}
\end{equation}



Both Anderson's ACT-R theory and Ausubel's meaningful learning theory imply the significance of connecting the new knowledge required to solve a problem to the existing concepts in the learner's knowledge base \update{(entailed in \Cref{eq:def4})}. We also account for the diversity of learners' motivation to study and incorporate the notion of improving an individual's learning objective (\Cref{eq:def3}). The hint generation strategy for a learner aiming to improve their answer accuracy would greatly differ from someone aiming to maximize the diversity of acquired knowledge in a learning session. 


\noindent\textbf{Extending the formal definition.} The given definition assumes a question-answering setup for the hint generation task, with an additional assumption of the availability of objective answers. We can modify this definition to accommodate for other hint generation settings as described below.

\begin{itemize}[itemsep=0pt]
\item For subjective questions, we should replace the correct answer $a$ with an evaluation rubric instead $\mathcal{R} : \mathcal{H} \rightarrow \mathbb{R}$.
\item For a hint generation system for the writing assistance task, we need to replace the question with the writing task description ($q \rightarrow T$) and ongoing dialogue with an interactive sequence of learner's writings ($\hat{a_i} \rightarrow \hat{w_i}$), and past hints ($\hat{h_i}$), $\mathcal{D}_q^l \rightarrow \mathcal{W}_T^l \mid \mathcal{W}_T^l \Leftrightarrow \{\hat{w_1}, \hat{h_1}, \hat{w_2}, \hat{h_2}, ...\}$, and the answer with a target rubric for writing evaluation $a \rightarrow \mathcal{R}$. 
\item For a multi-modal hint generation system, we can assume the atomic instances of the dialogue (namely, $q$, $a$, $h$, $\hat{a}$, and $\hat{h}$) are constituted of different modalities depending on the task specifications.
\end{itemize}

\subsection{Components of an Effective Automatic Hint Generation System} \label{sec:comp_tasks}
In this section, we briefly review some NLP research areas that can inform the design of an automatic hint generation system \update{(\Cref{fig:overview})} that aligns with the \update{proposed} formal definition introduced in Section \ref{sec:task_def}. 


\noindent\textbf{Question Answering.} 
Current hint generation systems are framed as scaffolding tools within a question answering (QA) setup, where learners are assessed on their ability to answer questions on a relevant topic. These systems can benefit from research in question answering in two ways.

Firstly, hint generation models should be equipped with reasoning abilities to answer questions. \citet{anderson1995cognitive} emphasizes the significance of an underlying production-rule model that helps to break down a learning goal into achievable subgoals in their cognitive tutoring framework, and we posit the QA systems today can form the foundation of this production-rule model. Datasets like \texttt{StrategyQA} \cite{geva2021did} and strategies like \texttt{Self-Ask} \cite{press2022measuring} and \texttt{Socratic Questioning} \cite{qi2023art} are great illustrations of the goal decomposition ability described in the ACT-R cognitive learning theory framework \cite{anderson1995cognitive}. 
\update{
We can adapt these computational question answering frameworks coupled with the answer assessment module (described below) to learn mapping $\textbf{H}: \{q, a, \mathcal{K}_{q\rightarrow a}, D^l_q\} \rightarrow \mathcal{H}$.}

Secondly, question answering systems are quite diverse and can solve complex questions spread across multiple modalities such as knowledge bases \cite{lan2022complex}, tables \cite{jin2022survey}, images \cite{srivastava2021visual, de2023visual}, and videos \cite{zhong2022video}, as well as explain the reasoning for the answer \cite{danilevsky2020survey, schwalbe2023comprehensive}. These question answering models can form the basis of hint generation systems that take into account incomplete solutions and the necessary reasoning steps to answer the question.

\noindent\textbf{Answer Assessment.}
Answer assessment plays a crucial role in evaluating and understanding learner responses, enabling the system to provide targeted feedback and adaptive guidance. Accurate answer assessment is foundational for standardizing the grading process, identifying misconceptions, tracking individual progress, and tailoring subsequent hints to address specific learning needs.
\update{
We can utilize an answer assessment module to identify the missing (or wrong) components of an ongoing answer ($\hat{a}_k$) juxtaposed to a reference answer ($a$) and provide appropriate hints to scaffold them towards a correct solution. We can also utilize the answer assessment module to aggregate the mistakes made by a learner over time to track their progress.
}

Prior work includes several answer evaluation strategies, varying from naive exact match and token overlap approaches to BERT-based semantic strategies for short form answers \cite{bulian2022tomayto}, explainable systems trained on LLM-distilled rationales \cite{li2023distilling}, semantic grouping based systems for batch grading \cite{chang2022towards} and multi-modal assessment frameworks to evaluate oral presentations \cite{liu2020personalized}. For a detailed overview of answer assessment strategies, we point the readers to the survey written by \citet{das2021automatic}.

\noindent\textbf{User Modeling.} User modeling is extensively explored within the recommendation systems research area, with the aim of adapting and customizing a service to a user's specific needs. User modeling holds the key to identifying the learner's interests and preferences within the hint generation framework (\Cref{eq:def4}), helping align the hints to the learner's prior knowledge (Section \ref{sec:semantics}) and personalize the feedback (Section \ref{sec:pragmatics}). 
\update{
Using a learner's past interaction data ($\mathcal{L}_l$), we can learn a preference function ($\mathcal{F}^l_{pref}$) based on the success of previous hints in helping the learner achieve their learning goal. We can measure this learning goal ($\mathcal{F}^l_{learning}$) using direct or indirect learner feedback from the user interaction. For instance, we can obtain learners' self-reported data about the effectiveness of a hint, or develop alternate indicators of success by observing the learner's behavior across several dimensions, such as the number of interactions, response initiation time, degree of change in response, and frequency of hints asked.
}

\citet{liu2022open}, for instance, utilizes long short-term memory (LSTM) cells \cite{hochreiter1997long} for knowledge estimation \cite{corbett1994knowledge} of student's current understanding based on their past responses for open-ended program synthesis. 
\update{
They utilize these time-varying student knowledge states to predict students' responses to programming problems, monitoring and analyzing their progress. We can adopt similar user modeling techniques to measure progress, adapt the hints to the learner's preferences, and design a curriculum (or suggest areas) for further improvement.}
We direct the readers to the comprehensive survey authored by \citet{he2023survey} for an extensive overview on user modeling.

\noindent\textbf{Dialogue Generation.} Intelligent tutoring systems are mediated by dialogue systems (or conversational agents), providing an interface between ITS and learners. Recent developments in retrieval augmented generation \cite{gao2023retrieval, 10.1145/3477495.3532682, li2022survey}, emotion-aware dialogue systems \cite{ma2020survey}, and enhanced language understanding capabilities in multi-turn dialogues \cite{zhang2021advances} can help improve the hint generation models, empowering them with abilities to better understand the learner's queries, understand and respond to complex domains, generate affective responses, and keep track of long conversation sessions for adaptive tutoring that would help link the generated hints to learner's prior knowledge (Section \ref{sec:semantics}). For further exploration, we invite readers to read the surveys written by \citet{ni2023recent}, \citet{deriu2021survey} and \citet{ma2020survey}.

\noindent\textbf{Question Generation.} Questions play a pivotal role in education, serving to help recall knowledge, test comprehension, and foster critical thinking. \citet{AlFaraby2023} classify neural question generation for educational purposes into three broad categories: (1) Question Generation from Reading Materials, (2) Word Problem Generation, and (3) Conversation Question Generation. We posit a similar use case for question generation to generate hints within conversations to clarify not just the learner's response but also their understanding and ability to present their ideas. We point the readers to surveys written by \citet{kurdi2020systematic}, \citet{zhang2021review}, \citet{das2021automatic} and \citet{pan2019recent} for a comprehensive overview.




\noindent\textbf{Modular structure of proposed approach.} \update{Although an ideal hint generation system would be able to satisfy all the criterion described in our formal definition (\Cref{eq:def1,eq:def2,eq:def3,eq:def4})---providing hints that are personalized to individual learners preferences and learning objectives---we acknowledge the limitations of technology at the time of this survey (\Cref{sec:future_directions}). Keeping that in mind, our proposed formal definition (\Cref{sec:task_def}) and components of hint generation system (\Cref{sec:comp_tasks}) suggests a modular structure, comprising of some essential components (such as \Cref{eq:def1,eq:def2} that form the crux of a hint generation system, or the question answering module that helps develop the reasoning to develop the hints) and some complementary modules (like \Cref{eq:def3,eq:def4} that helps personalize the hints to individual learners or question generation module that adapts particular hint giving strategies) to further enhance the quality of generated hints.
This paper proposes thus a long term roadmap for automatic hint generation systems that can showcase intermediate progress along the way. 
}



\subsection{Open Challenges for Effective Automatic Hint Generation Systems} \label{sec:future_directions}

In this section, we outline challenges and future directions for building more effective automatic hint generation systems that align with our formal definition. 

\noindent\textbf{Privacy-preserving self-evolving frameworks.} Current hint generation frameworks limit their applications to a fixed dataset or a pre-defined set of problems, which does not 
\update{
guarantee high quality performance in real-world applications. In order to make these frameworks more effective for the learners, we need to adopt self-evolving frameworks that can incorporate the learner's feedback and implicit preferences. These self-evolving frameworks should have the capability of identifying user feedback from an ongoing conversation ($D^l_q$), gathering implicit preferences ($\mathcal{F}^l_{pref}$) from past learning interactions ($\mathcal{L}_l$), and adapting their hint generation process to an individual learner while respecting their right to privacy.} 

\update{Although there is prior work on incorporating human-feedback to improve generation quality (\textit{e.g.,} prompt optimization \cite{chang2024efficient, sahoo2024systematic}, reinforcement learning with human feedback \cite{kaufmann2023survey, casper2023open}), incorporating these user-modeling aspects in a privacy-preserving manner remains an active area of research \cite{miranda2024preserving}. Differential privacy \cite{zhao2022survey, yang2023local} and federated learning \cite{li2021survey} could be relevant research avenues to help create these self-evolving frameworks with a responsible approach. For an in-depth review of the state of preserving privacy in LLMs, readers are encouraged to consult the survey by \citet{miranda2024preserving}.
}

\noindent\textbf{Diverse domain exploration.} Most efforts within the hint generation space are limited to programming, language acquisition or mathematics, similar to the domain trends in intelligent tutoring systems \cite{mousavinasab2021intelligent}. However, education and tutoring often involve other subjects within natural sciences (\textit{e.g.,} physics, chemistry, biology, earth sciences), social sciences (\textit{e.g.,} history, civics, geography, law), and learning beyond educational institutions (\textit{e.g.,} educating patients about their health conditions effectively \cite{gupta-etal-2020-heart}). Tackling the hint generation problem in these subjects raises several challenges, such as the evaluation of subjective long-form answers, the need for domain knowledge and enhanced reasoning capabilities \cite{10.1162/coli_a_00511,huang2022towards,zhang2023igniting}. However, it simultaneously opens up new opportunities to expand the capabilities of hint generation systems, including the development of smaller models that combine the power of pre-trained language models with the power of adapted knowledge models (e.g., COMET \cite{bosselut-etal-2019-comet}).  

\update{Efforts in question answering can also help provide seed datasets and data annotation strategies to expand the hint generation resources. The literature is rich in education-related datasets ranging from generic datasets like \texttt{RACE} \cite{lai2017race}, \texttt{LearningQ} \cite{chen2018learningq}, \texttt{TQA} \cite{kembhavi2017you} \textit{etc.} to subject-specific datasets like \texttt{SciQ} \cite{welbl2017crowdsourcing}, \texttt{ScienceQA} \cite{lu2022learn}, \texttt{SituatedQA} \cite{zhang2021situatedqa}, \texttt{GSM8K} \cite{cobbe2021training}, \texttt{BioASQ} \cite{tsatsaronis2015overview}, \texttt{CORD-19} \cite{wang2020cord}, \texttt{PubMedQA} \cite{jin2019pubmedqa}. 
}
To extend these datasets  
we envision a collaborative approach between the ML practitioners and expert tutors to explore the utilization of hints in classroom or one-on-one tutoring setting across different domains. 





\noindent\textbf{Multi-lingual and multi-cultural aspects.} Prior studies have found a correlation between linguistic and cultural diversity and capabilities for innovation \cite{hofstra2020diversity, evans2009myth}. However, we found that the majority of research at the intersection of natural language processing and learning sciences is limited to the English language, either as the mode of education or a subject for language acquisition. 
\update{
Although utilizing multi-lingual large language models (mLLMs) for hint generation can potentially help incorporate the linguistic aspects of the learning experience, providing culturally-aware hints still remains a challenging task. \citet{liu2024multilingual} reveals that while mLLMs are aware of cultural proverbs, they struggle to reason with figurative proverbs and sayings. Building benchmarks to evaluate cultural awareness of generation models such as \texttt{MAPS} \cite{liu2024multilingual}, \texttt{PARIKSHA} \cite{liu2024multilingual}, \texttt{CultureAtlas} \cite{fung2024massively} and \texttt{CUNIT} \cite{li2024well} is an essential first step towards identifying the shortcomings of hint generation frameworks. Moving forward, we need to explore the hint generation capabilities of mLLMs and develop culturally-aware systems that accommodate the education for non-English learners, and sustain the diverse cultural and linguistic heritage \cite{bernard1992preserving, soto1999preserving}.
}


\noindent\textbf{Multi-modal elements.} 
Prior qualitative research has established gains from complementing education with additional modalities like maps, diagrams, and multi-media content \cite{winn1991learning, swidan2019role, tippett2016recent, kapi2017multimedia, abdulrahaman2020multimedia}. 
\update{
Research in intelligent tutoring systems has also explored incorporating certain gamifying elements like badges, leaderboards, narratives, and virtual currency 
to keep the learners more engaged and motivated \cite{gonzalez2014gamification, ramadhan2024combining}.
}
We believe incorporating these cues into a hint generation system would improve the students' memory, understanding, and overall learning experience to help create a holistic educational tool \cite{pourkamali2021effect}.



\noindent\textbf{Affective systems.} 
Affective aspects are often neglected when building intelligent tutoring systems and hint generation systems \cite{hasan2020transition}. However, incorporating them into the education pipeline is key for personality development,  encouragement and improving self-motivation \cite{jimenez2018affective}. \citet{jimenez2018affective} shows that using affective feedback has a positive impact on students facing learning challenges.
Thus, affective hint generation systems that take into consideration the emotional state of the learners when providing feedback is an important area of study. 

\update{
In recent years, there has been increased interest in building dialogue agents that can adapt to the emotional state of the user in an ongoing dialogue setting. \citet{haydarov2023affective} develops a large-scale multi-modal benchmark for visually grounded emotional reasoning-based conversations that use visual stimuli to stir a conversation to test out the emotional reasoning capabilities of multi-modal systems. \citet{li2023know}, on the other hand, proposed a future emotion state prediction framework in spoken dialogue systems to predict the future affective reactions of users based on the ongoing conversation. 
Resources and findings from the ongoing research in affective dialogue systems could be leveraged and advanced to develop more adaptive and emotion-aware tutoring systems. We point the readers to \citet{ma2020survey}, \citet{raamkumar2022empathetic}, and \citet{zhang2024affective} for a comprehensive review of affective dialogue systems.
}


\noindent\textbf{Accessible systems.} Another fruitful and challenging area of research is to develop accesible hint generation systems. 
For people with neurodevelopmental disorders (\textit{e.g.,} attention-deficit/hyperactivity disorder (ADHD) and autism spectrum disorder (ASD)) and learning disabilities (\textit{e.g.,}  dyslexia and dyscalculia), one can modify the hint generation systems to have: (1) \textit{simplified text}: using plain language, and avoid using jargons and complex terminology \cite{vstajner2021automatic}, (2) \textit{multi-sensory supports}: leveraging a combination of visual, auditory, and kinesthetic modalities to present hints in multiple formats \cite{vezzoli2017dyslexia, gori2014perceptual}, (3) \textit{interactive elements}: incorporating interactive elements to engage learners to explore concepts in a hands-on manner \cite{garcia2018interactive}, and (4) \textit{predictable routine}: establishing a consistent routine for delivering hints consistently can helps learners feel more comfortable and confident \cite{love2012effects}. 

\noindent\textbf{Evaluation metrics.} Evaluation of generated hints is a non-trivial task, depending on multiple factors to determine the quality and the success of a hint. Prior work determines the success of a hint generation system by the learner's abilities in producing the reference solution (\Cref{eq:def2}) \cite{jatowt2023automatic}. However, this evaluation framework does not penalize the generated hints if they make the problem too simple, and also does not take into account the individual learner's preferences and learning objectives. Therefore, we need to build human-centered evaluation frameworks \cite{lee2022evaluating} that can help measure factors beyond the learner's answering capabilities, such as learner's ownership over the learning process, long-term capabilities of generated knowledge, motivation and enjoyment they receive during their interactions. 

\subsection{Ethical considerations} \label{sec:ethics}

The integration of NLP within educational settings raises distinct concerns, such as the impact on pedagogical approaches, the dynamics of teacher-student interaction, and learner agency \citep{holstein2019designing}. The adoption of NLP technologies in the classroom implements a particular theory of teaching and learning, and these values must be made explicit \cite{blodgett2021risks}. How does introducing a new tool reconfigure the dynamics of the teacher-student relationship? Here, it would be crucial to avoid the solutionism trap, define the boundaries of where the system is useful, and ensure that the intention is to augment educators' workflows instead of substituting them \citep{remian2019augmenting}. Researchers and practitioners must also attend to the longer-term impacts of engaging with young individuals during a formative period \cite{holmes2021ethics}. Below, we outline various ethical considerations, including data privacy and consent, bias and fairness, and effects on language variation \cite{schneider2022multilingualism}, and offer strategies to address these concerns. 

One of the biggest sets of ethical considerations relates to the use of student and teacher information \citep{nguyen2023ethical}. Given the sensitive nature of educational data, it will be important to set up privacy measures and enable informed consent. Students' information beyond individual responses to questions may need to be tracked to provide an effective learning experience \citep{kerr2020ethical}. However, this opens up the possibility of surveillance and misuse, jeopardizing learners' trust and autonomy \cite{regan2019education}. It would be critical to promote data literacy among educators and learners (\textit{e.g.,} through workshops) to enable them to minimize the risk of their participation \citep{kerr2020ethical}. The issue of data ownership raises questions about who holds control over the information collected through education platforms.

We must collectively explore the broader implications of integrating NLP in education on representativeness and equity and exacerbating systemic inequalities \citep{weidinger2022taxonomy}. There is a risk that the hints generated by NLP models may not adequately reflect the diverse backgrounds and lived experiences of students \citep{dixon2020racializing} and potentially perpetuate harmful stereotypes about different identities \citep{dev2021measures}. Prior work has demonstrated the various forms of `bias' in NLP systems \citep{blodgett2020language}, which may contribute to the construction of language hierarchies and limiting language variation \citep{schneider2022multilingualism}. To promote inclusive design and mitigate these ethical considerations, it is essential to understand how power, privilege, and resources are redistributed as a result of introducing AI in the classroom. Is there a possibility of diminishing quality education for marginalized and under-resourced groups \citep{remian2019augmenting}? We must take a community-collaborative approach to understand how to design justice-oriented and accountable systems \citep{madaio2022beyond} where learners can truly benefit from hint generation systems.

\section{Conclusion}


In this paper, we consolidate prior research in hint generation, bridging the gap between research in education and cognitive science, and research in AI and natural language processing. Based on our findings, we propose a roadmap for the future research in hint generation, where we provide a rethinking of the formal task definition, a brief review of research areas that can inform the design of future systems, open challenges for effective hint generation systems, and the ethical considerations.
Although hint generation has a long history dating back over three decades \cite{hume1996hinting}, recent advances in natural language processing could serve useful for future hint generation systems.
Beyond education, hint generation is also an excellent atomic task to measure a system's ability to personalize content to user needs and requirements. We invite researchers to foster a community, develop new benchmarks, create shared tasks and workshops for automatic hint generation.






\section*{Limitations}

Due to the rich and diverse literature on intelligent tutoring systems, 
we limit our survey to research directly relevant to hint generation and do not cover other types of \update{learning-related} feedback. 



\bibliography{references/general, references/tutoring, references/education, references/ai, references/ethics}
\bibliographystyle{acl_natbib}


\end{document}